\documentclass[lettersize,journal]{IEEEtran}
\usepackage{amsmath,amsfonts}
\usepackage{amssymb}
\usepackage{algorithmic}
\usepackage{array}
\usepackage[caption=false,font=normalsize,labelfont=sf,textfont=sf]{subfig}
\usepackage{textcomp}
\usepackage{stfloats}
\usepackage{url}
\usepackage{verbatim}
\usepackage{graphicx}
\usepackage{graphicx}
\usepackage{epstopdf}
\usepackage{xcolor}
\usepackage{booktabs}
\usepackage{array}
\usepackage{multirow}
\usepackage{threeparttable}
\usepackage{amsmath,amsfonts}
\usepackage{algorithmic}
\usepackage{algorithm}
\usepackage{array}
\usepackage[caption=false,font=normalsize,labelfont=sf,textfont=sf]{subfig}
\usepackage{textcomp}
\usepackage{stfloats}
\usepackage{url}
\usepackage{verbatim}
\usepackage{graphicx}
\usepackage{pdfpages}
\usepackage{float}  
\usepackage{booktabs} 
\usepackage{cite}
\usepackage{makecell}
\usepackage{tabu}                     
\usepackage{multirow}                 
\usepackage{multicol}                 
\usepackage{multirow}                
\usepackage{float}                    
\usepackage{makecell}                 
\usepackage{booktabs}                 
\usepackage{threeparttable}
\usepackage{tikz}

\def\BibTeX{{\rm B\kern-.05em{\sc i\kern-.025em b}\kern-.08em
    T\kern-.1667em\lower.7ex\hbox{E}\kern-.125emX}}
\usepackage{balance}
\begin{document}
\title{Wavelet-Guided Dual-Frequency Encoding for Remote Sensing Change Detection}
\author{Xiaoyang Zhang, Guodong Fan, Guang-Yong Chen, ~\IEEEmembership{Member,~IEEE}, Zhen Hua, Jinjiang Li, \\ Min Gan, ~\IEEEmembership{Senior Member,~IEEE}, C. L. Philip Chen, ~\IEEEmembership{Life Fellow, IEEE}
\thanks{This research was supported by the National Natural Science Foundation of China (62301105, 61772319, 62272281, 62002200, 62202268), Shandong Natural Science Foundation of China (ZR2020QF012 and ZR2021MF068), Yantai science and technology innovation development plan(2022JCYJ031).(Corresponding author: Guodong Fan.)}
\thanks{
X. Zhang, G. Fan, Z. Hua and J. Li are with School of Computer Science and Technology,
Shandong Technology and Business University, Yantai 264005, China}
\thanks{Guang-Yong Chen is with the College of Computer and Data Science, Fuzhou University, Fuzhou 350116, China, and also with the Fujian Key Laboratory of Network Computing and Intelligent Information Processing, the Key Laboratory of Intelligent Metro of Universities in Fujian, and the Engineering Research Center of Big Data Intelligence, Ministry of Education, Fuzhou 350108, China.}
\thanks{Min Gan is with the College of Computer Science and Technology, Qingdao University, Qingdao 266071, China. (aganmin@aliyun.com)}
\thanks{C. L. Philip Chen is with the School of Computer Science and Engineering, South China University of Technology, Guangzhou 510641, China. (philip.chen@ieee.org)}
}
\markboth{Journal of \LaTeX\ Class Files,~Vol.~18, No.~9, September~2020}%
{How to Use the IEEEtran \LaTeX \ Templates}

\maketitle

\begin{abstract}
Change detection in remote sensing imagery plays a vital role in various engineering applications, such as natural disaster monitoring, urban expansion tracking, and infrastructure management. Despite the remarkable progress of deep learning in recent years, most existing methods still rely on spatial-domain modeling, where the limited diversity of feature representations hinders the detection of subtle change regions. We observe that frequency-domain feature modeling—particularly in the wavelet domain—can amplify fine-grained differences in frequency components, enhancing the perception of edge changes that are challenging to capture in the spatial domain. Thus, we propose a method called Wavelet-Guided Dual-Frequency Encoding (WGDF). Specifically, we first apply Discrete Wavelet Transform (DWT) to decompose the input images into high-frequency and low-frequency components, which are used to model local details and global structures, respectively. In the high-frequency branch, we design a Dual-Frequency Feature Enhancement (DFFE) module to strengthen edge detail representation and introduce a Frequency-Domain Interactive Difference (FDID) module to enhance the modeling of fine-grained changes. In the low-frequency branch, we exploit Transformers to capture global semantic relationships and employ a Progressive Contextual Difference Module (PCDM) to progressively refine change regions, enabling precise structural semantic characterization. Finally, the high- and low-frequency features are synergistically fused to unify local sensitivity with global discriminability. Extensive experiments on multiple remote sensing datasets demonstrate that WGDF significantly alleviates edge ambiguity and achieves superior detection accuracy and robustness compared to state-of-the-art methods. The code will be available at https://github.com/boshizhang123/WGDF.
\end{abstract}


\section{Introduction}
\IEEEPARstart{R}{emote} sensing imagery has demonstrated significant value in various application domains, including urban planning, land use analysis, and disaster assessment \cite{1}, \cite{2}, \cite{3}. Building detection, as a critical component of change detection tasks, plays a vital role in improving the 
\begin{figure}[t]
\centering
\includegraphics[scale=0.445]{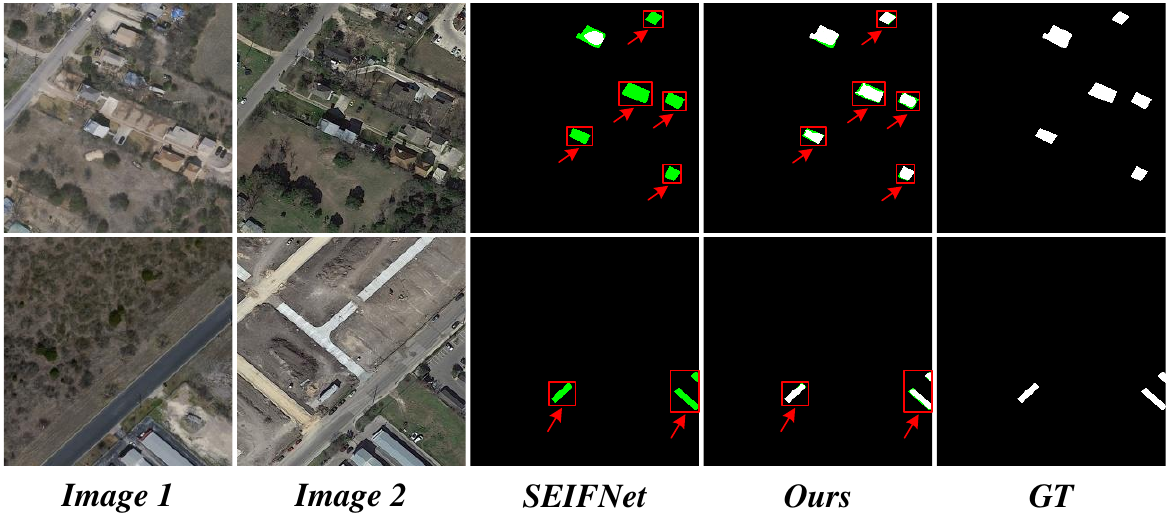}
\caption{Compared with the spatial-domain method SEIFNet, ours method delivers more accurate detection results in representative small-change regions (highlighted with red boxes, where green denotes undetected regions), demonstrating its superior capability in capturing subtle structural variations through high-frequency wavelet representations.}
\label{figure}
\end{figure}
performance of downstream applications such as object recognition and land cover classification \cite{4}, \cite{5}. However, the complex and diverse background textures, along with inconsistent spatial resolutions in remote sensing images, present considerable challenges for accurate building edge detection. 

Traditional change detection methods for building detection mainly fall into algebra-based and transformation-based categories \cite{6}. Algebra-based approaches, such as Change Vector Analysis (CVA) \cite{7} and Spectral Angle Mapping (SAM) \cite{8}, extract change information using pixel-level statistics or algebraic operations. Transformation-based methods like Principal Component Analysis (PCA) \cite{9}, Independent Component Analysis (ICA) \cite{10}, and Linear Discriminant Analysis (LDA) \cite{11} enhance change region separability by projecting data into discriminative feature spaces. However, these hand-crafted feature methods struggle with fine-grained building edge detection in complex backgrounds, especially under heterogeneous surfaces like vegetation and roads. Their limited capability in detailed feature extraction and global context modeling restricts performance, highlighting the need for more adaptive and intelligent feature modeling to accurately capture building edges in challenging scenarios.

Recent advances in deep learning have significantly enhanced change detection tasks \cite{12}. CNNs extract multi-scale hierarchical features that improve discrimination between buildings and complex backgrounds. Early single-stream fusion methods like FC-EF \cite{13} concatenate bi-temporal images for end-to-end detection but lack explicit temporal correlation modeling, limiting localization accuracy in complex scenes \cite{14}. Dual-stream Siamese architectures address this by independently extracting features from each temporal image and modeling their differences, improving building edge detection \cite{15}, \cite{16}. Attention mechanisms further enhance focus on critical features, dynamically adjusting weights to better separate buildings from backgrounds \cite{17},\cite{18}, \cite{19},\cite{20}. However, CNNs’ limited receptive field restricts their ability to capture long-range spatial dependencies, affecting performance amid complex backgrounds and subtle edge variations \cite{21}. To overcome this, Transformer-based models leveraging self-attention have been introduced for change and edge detection \cite{22}, \cite{23}, excelling at modeling global semantic context. However, in real-world remote sensing applications, accurate detection often hinges on the ability to sensitively capture subtle structural changes. Existing methods generally underperform in this regard, particularly struggling to distinguish slight changes from complex backgrounds. As illustrated in Fig. 1, conventional spatial-domain models fail to effectively detect the minor changes marked by red boxes—details that are of significant importance in many practical scenarios. To address this issue, we further extend our research perspective to frequency-domain analysis.

In domains such as medical image processing, texture analysis, and image compression, Discrete Wavelet Transform (DWT) has been widely adopted due to its superior multi-scale decomposition capability and spatial-frequency localized analysis. Compared to conventional Fourier transforms, DWT preserves both spatial and frequency information across multiple scales, demonstrating remarkable effectiveness in capturing local structures such as edges and textures. This property is particularly important for visual tasks characterized by complex structures and non-uniform information distribution. Given that remote sensing change detection also faces challenges such as weak local variations and significant background interference, the frequency-domain representation offered by DWT holds strong theoretical transferability to this task. However, although some studies have attempted to introduce frequency-domain operations into intermediate network layers \cite{24},\cite{25}, these methods essentially remain post-processing of spatial features and fail to decouple feature representations at the source. More importantly, the direct use of DWT at the input stage to explicitly separate low- and high-frequency components, and the subsequent design of a dual-branch architecture tailored to this property, remain largely unexplored.

To fill this gap, we propose a framework—Wavelet-Guided Dual-Frequency Encoding (WGDF). Unlike conventional methods, WGDF decomposes bi-temporal remote sensing images into low- and high-frequency components at the input, alleviating feature entanglement early on. A dual-branch architecture exploits frequency-domain complementarity: the high-frequency branch uses a Dual-Frequency Feature Enhancement (DFFE) module to refine fine-grained building edges and a Frequency-Domain Interactive Difference (FDID) module to capture temporal high-frequency variations for precise edge change detection. The low-frequency branch employs a Transformer to model global semantic relations, complemented by a Progressive Contextual Difference Module (PCDM) for progressive semantic change refinement. This collaborative dual-branch design enables WGDF to achieve more accurate and robust building edge detection in complex backgrounds, presenting a paradigm for high-precision remote sensing change detection.

Overall the main contributions of this paper are as follows:
\begin{itemize}

\item We thoroughly investigate the frequency-domain characteristics of remote sensing images and propose a change detection method, Wavelet-Guided Dual-Frequency Encoding (WGDF). This method significantly alleviates the edge confusion issue in complex backgrounds through the co-optimization of high-frequency and low-frequency branches.

\item To address the challenge of detail extraction from high-frequency features, we propose an iteratively optimized Dual-Frequency Feature Enhancement Module (DFFE), which significantly enhances the ability to depict building edges in complex backgrounds by iteratively improving the expression of high-frequency features.

\item To accurately capture changes in high-frequency features, we propose the Frequency-Domain Interactive Difference Module (FDID), which enhances the detection accuracy of edge-change regions through fine-grained modeling of high-frequency features.

\item To fully capture variations in low-frequency features, we propose the Progressive Contextual Difference Module (PCDM), which enhances the modeling of global contextual information by gradually extracting semantic differences between dual-temporal images.

\end{itemize}

\section{Related work}
\noindent 

\subsection{Deep Learning Based Change Detection Methods}

Change detection models typically employ CNNs or Transformers to enhance semantic feature representation via optimized architectures, loss functions, and attention mechanisms \cite{26}. Early methods were dominated by Siamese networks due to their natural suitability for processing bi-temporal remote sensing images \cite{26}, and this framework remains widely used. FCN-based models like FC-EF and FC-Siam-Conc \cite{13} process dual-temporal images without preprocessing. DTCDSCN \cite{27} integrates channel and spatial attention for refined detail extraction, while STANet \cite{28} introduces spatio-temporal attention to capture long-range dependencies. SNUNet \cite{29} fuses shallow and deep features through dense connections to improve robustness and accuracy. Transformers were introduced to change detection by BIT \cite{17}, leveraging their broad receptive fields to model spatio-temporal context. ChangeFormer \cite{30} combines a hierarchical Transformer encoder with a perceptual decoder to capture multiscale changes. ICIFNet \cite{31} addresses misalignment via intra- and inter-scale feature fusion, and ChangStar \cite{32} enhances generalization by exchanging temporal inputs. DMINet \cite{33} proposes a cross-period attention module for deep intra- and inter-layer coupling, while ELGC-Net \cite{34} combines local-global context aggregation with attention to improve change region estimation. Despite these advances, challenges remain in complex backgrounds. High-level convolution features can obscure narrow boundaries (e.g., building edges), weakening edge detection. SMNet \cite{1} enhances change detection by fusing multi-level features using a hybrid semantic-guided Mamba and efficient RWKV architecture. Furthermore, most methods prioritize spatial features and underutilize frequency-domain details, limiting the precision of change region modeling.

Unlike previous methods, we significantly enhance the model’s ability to capture edge details and model global semantics in complex backgrounds by incorporating frequency-domain features into the change detection task and jointly optimizing high-frequency and low-frequency branches. The high-frequency branch focuses on detailed edge feature extraction, while the low-frequency branch captures global background information. This design effectively addresses the issues of edge boundaries being easily overlooked and the underutilization of frequency-domain features in existing approaches.

\subsection{Frequency-Domain Based Learning of Visual Representations}

With the advancement of deep learning, visual representation learning has become central in computer vision. While traditional methods rely on spatial-domain convolutions to extract local and global features, frequency-domain representations have drawn increasing attention due to their advantages in texture encoding, detail enhancement, and noise suppression \cite{35}, \cite{36}. By decomposing images into different frequency components, frequency-domain analysis enables independent modeling of global and local features, offering benefits for complex visual tasks \cite{37}. In Remote Sensing Change Detection (RSCD), frequency-domain approaches enhance robustness against noise. HFA-Net \cite{38} applies high-frequency attention for precise localization, while LPEM \cite{39} preserves high-frequency details during decoding. ESACD \cite{40} improves feature representation by separately fusing low- and high-frequency information. Transformer-based models like DCAT \cite{41} integrate dual cross-attention to extract frequency-specific features, and Gao et al. \cite{42} enhance region isolation with frequency analysis and stochastic mapping. ASGF \cite{43} combines spatial and frequency features with pseudo-label optimization for unsupervised detection. Wavelet transform, as an efficient frequency-domain tool, is widely adopted \cite{44}. Liu et al. \cite{45} emphasize the role of low- and high-frequency components for semantic modeling and fine details in video segmentation. Yang et al. \cite{46} use wavelet attention for image classification, while Cheng et al. \cite{47} decouple DWT-based frequency features to better capture both semantic and detail information.

Unlike previous methods that apply wavelet transform during intermediate processing, our approach introduces a use of discrete wavelet transform (DWT) at the input stage to directly decompose images into low- and high-frequency components. This design enables full extraction of frequency-domain information from raw data and provides a strong foundation for feature modeling. The low-frequency branch captures global semantics, while the high-frequency branch enhances details. Their joint optimization significantly boosts change detection robustness and accuracy in complex scenes. Compared with conventional designs, our method more effectively exploits wavelet transform’s potential, offering a promising direction for RSCD.
\begin{figure}[t]
\centering
\includegraphics[scale=0.4]{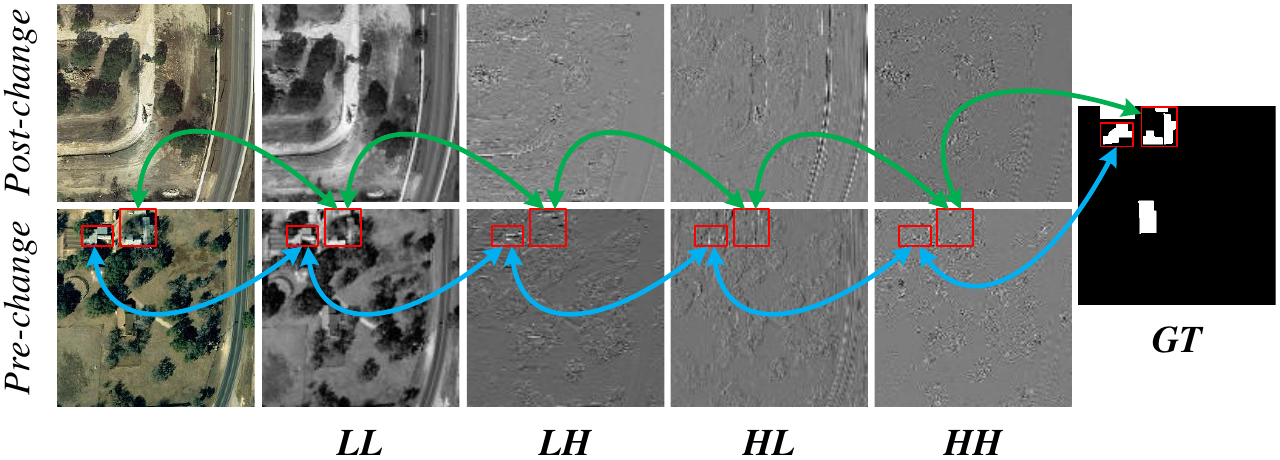}
\caption{An example of frequency-domain decomposition using DWT. The middle four columns show the low-frequency (LL) and high-frequency (LH, HL, HH) components obtained after DWT decomposition. The region enclosed by the red box corresponds to a small target region that is more difficult to detect, with the green and blue arrows highlighting the performance of this region across different frequency-domain features.}
\label{figure}
\end{figure}

\section{Methodology}
\label{section3}

\subsection{Motivation}

In remote sensing change detection, targets such as buildings often exhibit subtle and localized structural variations. These fine-grained changes are particularly prone to being overwhelmed by abundant unstructured background information in scenarios characterized by similar surface textures, complex backgrounds, or small target scales, thereby posing significant challenges to accurate detection. Most existing methods rely heavily on spatial-domain modeling; however, the inherent limitations of the receptive field make it difficult to simultaneously capture global semantic context and local details. Moreover, the strong coupling between spatial semantics and frequency-related details often leads to redundancy and interference during feature fusion, ultimately resulting in ambiguous or degraded decision-making.

This situation raises a fundamental question: how can we break the bottleneck caused by the entanglement of semantics and details at the perceptual level to achieve more precise detection of subtle structural changes? To address this, we propose extending the analytical perspective beyond the spatial domain into the frequency domain, leveraging the Discrete Wavelet Transform (DWT) to enable effective multiscale and localized frequency decomposition. Owing to its inherent spatial-frequency localization properties, DWT decomposes an image into low-frequency components (LL) that primarily encode semantic content and high-frequency sub-bands (LH, HL, HH) that emphasize structural and edge details. As illustrated in Fig. 2, Changes in small buildings exhibit significantly enhanced responses in the high-frequency sub-bands, revealing that frequency-domain features are inherently more sensitive to subtle changes than their spatial-domain counterparts.

More importantly, frequency-domain decomposition not only reveals the latent value of fine-grained details in change detection but also offers a natural pathway for decoupling semantic and structural information. Unlike conventional approaches that typically integrate frequency-domain modules into intermediate network layers, we take a step by applying DWT at the input stage. This early-stage transformation enables an explicit separation between semantic structures and fine details, upon which we design a differentiated dual-branch architecture to mitigate feature redundancy 
\begin{figure*}[htbp]
\centering
\includegraphics[scale=0.545]{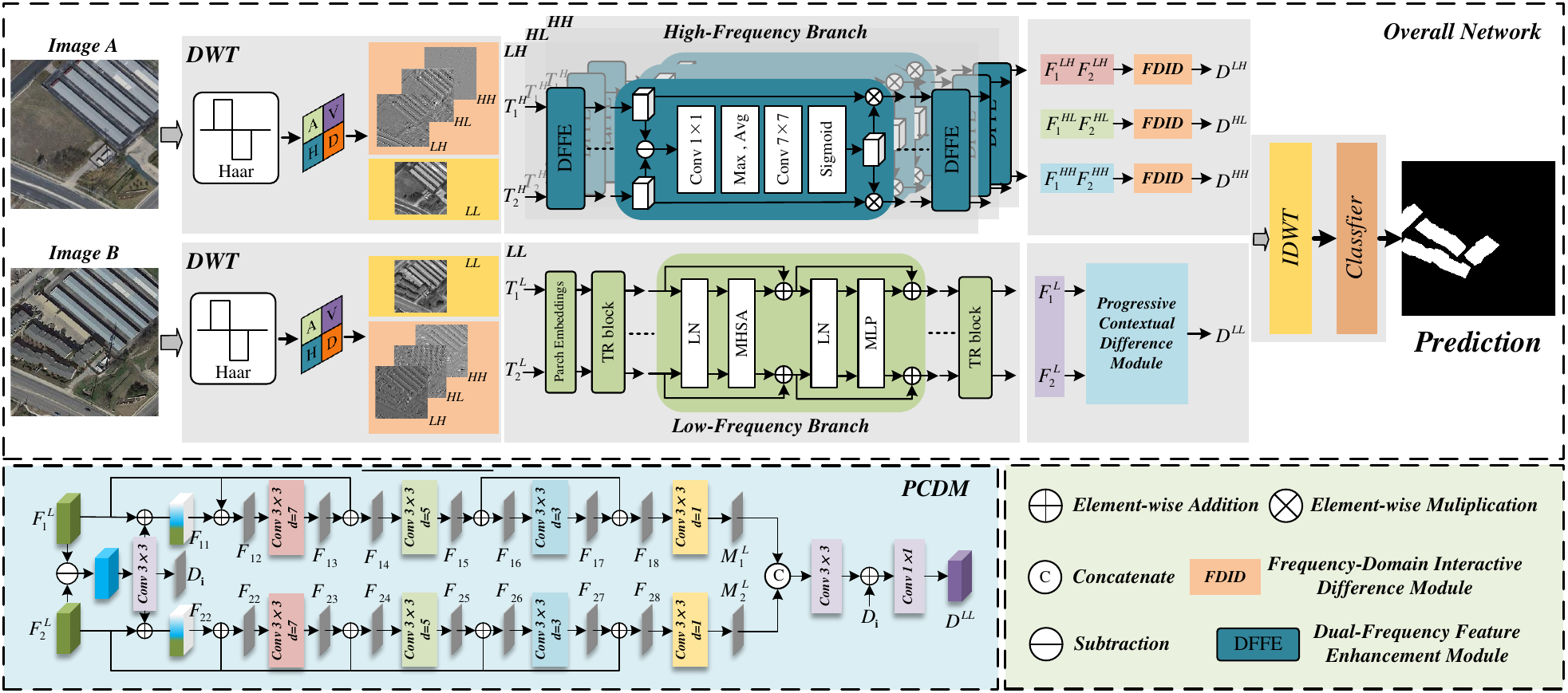}
\caption{The overall network structure of WGDF. The diachronic image is decomposed into low-frequency (LL) and high-frequency (LH, HL, HH) components using DWT (Haar). The high-frequency branch enhances edge feature extraction through DFFE and FDID, while the low-frequency branch employs the Transformer module and PCDM to capture global semantic changes. Finally, the change region prediction is obtained through IDWT and classifiers.}
\label{figure}
\end{figure*}
and interference from the outset. This spatial-frequency collaborative modeling strategy establishes a solid theoretical foundation for the WTDF framework, enhancing the perception of subtle changes while offering a direction for overcoming the limitations of purely spatial-domain methods.

\subsection{Network Architecture}

As shown in Fig. 3, the WGDF network adopts a dual-branch architecture for high and low-frequency features. The network first performs frequency-domain decomposition of bi-temporal remote sensing images using DWT, obtaining low-frequency (LL) and high-frequency (LH, HL, HH) components. The low-frequency branch introduces a stacked Transformer structure to model global semantic dependencies, combined with the PCDM to progressively extract low-frequency semantic changes, enhancing change perception in complex backgrounds. The high-frequency branch utilizes the DFFE to model features of the three high-frequency sub-bands and employs the FDID to capture fine-grained change differences, strengthening the representation of edge and structural changes. Subsequently, a fusion strategy for high- and low-frequency features enables complementary modeling of semantics and details. Finally, the features are restored to the spatial domain via IDWT, and the 
\begin{algorithm}[!ht]
\small 
\renewcommand{\algorithmicrequire}{\textbf{Input:}}
\renewcommand{\algorithmicensure}{\textbf{Output:}}
\caption{Wavelet-Guided Dual-Frequency Encoding for Remote Sensing Fine-Grained Change Detection}
\label{power}
\begin{algorithmic}[1]
\REQUIRE 
$Image A=x_i, i \in [1, n]$;\\
$Image B=y_i, i \in [1, n]$;\\
$GT=G_i, i \in [1, n]$;
\ENSURE Prediction: $Y$;
\STATE \textbf{DWT Decomposition:}
\STATE $DWT(x_i)=[HH_1,HL_1,LH_1,LL_1]$
\STATE $DWT(y_i)=[HH_2,HL_2,LH_2,LL_2]$
\STATE \textbf{High-Frequency Branch Feature Extraction:}
\STATE $(T_1^H, T_2^H) = [(HH_1,HL_1,LH_1),(HH_2,HL_2,LH_2)]$
\FOR{$j=2,3$}
\STATE $(T_j^H, T_j^H) = DFFE(T_1^H, T_2^H)$
\ENDFOR
\STATE \textbf{Low-Frequency Branch Feature Extraction:}
\STATE $(T_1^L, T_2^L) = [(LL_1),(LL_2)]$
\FOR{$j=2,3$}
\STATE $(T_j^L, T_j^L) = TR\ block(T_1^L, T_2^L)$
\ENDFOR
\STATE \textbf{High-Frequency Difference Feature Computation:}
\FOR{$m=HH,HL,LH$}
\STATE $D^m = FDID(F_1^m, F_2^m)$
\ENDFOR
\STATE \textbf{Low-Frequency Difference Feature Computation:}
\STATE $D^{LL} = PCDM(F_1^{LL}, F_2^{LL})$
\STATE \textbf{Image Reconstruction:}
\STATE $Change\ map=IDWT(D^{HH},D^{HL},D^{LH},D^{LL})$
\STATE $Y=Classfier(Change\ map)$
\STATE \textbf{return} Prediction: $Y$
\end{algorithmic}
\end{algorithm}
change detection results are output by the classifier. Algorithm 1 illustrates the WGDF workflow, clarifying the model’s design logic and operational process.

\subsection{Dual-Frequency Feature Enhancement Module (DFFE)}

High-frequency components contain rich edge and detail information but are easily affected by complex background interference such as textures and noise. To address this, we propose the Dual-Frequency Feature Enhancement Module (DFFE), as illustrated in Fig. 3, which progressively refines the high-frequency subbands (HH, HL, LH) through multi-scale modeling. Each subband is processed independently to accurately extract edge and detail features, enhancing the model’s robustness and precision in capturing fine-grained changes in complex scenes.

The input to this module includes high-frequency features $T_1^H$ and $T_2^H$ from bi-temporal remote sensing images. A difference operation first yields an initial high-frequency change map $T^D$, capturing key variation cues. To reduce complexity and preserve semantic richness, $T^D$ is processed by a $Conv_{1\times1}$ to compress channels while enhancing local feature representation. To further capture global dependencies, a channel attention mechanism is introduced. Specifically, spatial max-pooling and average-pooling are applied to $T^D$ to generate two complementary descriptors, $F_{max}$ and $F_{avg}$, which are concatenated and passed through a $Conv_{7\times7}$ layer. This operation captures inter-channel correlations, and the resulting feature map is normalized via Sigmoid to produce the channel-wise attention weights $W$. Finally, $W$ is used to adaptively recalibrate the original high-frequency inputs $T_1^H$ and $T_2^H$, selectively enhancing discriminative responses in change regions while suppressing redundant or noisy background information. This attention-guided refinement enables the network to generate more robust and informative high-frequency features for subsequent change detection tasks. The overall operation can be formulated as follows:
\begin
{equation}
(F_{max},F_{avg})=(Max,Avg)(Conv_{1\times1}(T^D)),
\end
{equation}  
\begin
{equation}
F_{i}^{H}=\sigma(Conv_{7\times7}(F_{max}+F_{avg}))\odot T_{i}^{H} (i=1,2),
\end
{equation} 
where $\sigma$ denotes the Sigmoid operation and $\odot$ represents the element-wise Hadamard product.

To capture deeper high-frequency variations, the DFFE module is stacked three times sequentially. Each stacked module progressively refines the high-frequency features layer by layer. Ablation experiments show that stacking three DFFE modules achieves the best performance. This multi-level feature enhancement effectively models high-frequency changes in complex backgrounds, significantly improving detection accuracy and robustness.

\subsection{Frequency-Domain Interactive Difference Module (FDID)}

In remote sensing change detection, high-frequency features encode essential edge and detail information vital for identifying subtle changes. However, complex backgrounds—such as texture variations, noise, and illumination changes—often degrade these features, leading to blurred or missing edge representations. To address this challenge, we propose the Frequency-Domain Interactive Difference Module (FDID), illustrated in Fig. 4, which is specifically designed to extract precise difference features from the high-frequency components of bi-temporal remote sensing images, thereby enabling accurate and robust detection of subtle changes under complex background conditions.

Specifically, the FDID module receives the high-frequency features from bi-temporal remote sensing images and computes the initial difference feature D. To capture multi-scale change 
\begin{figure}[htbp]
\centering
\includegraphics[scale=0.408]{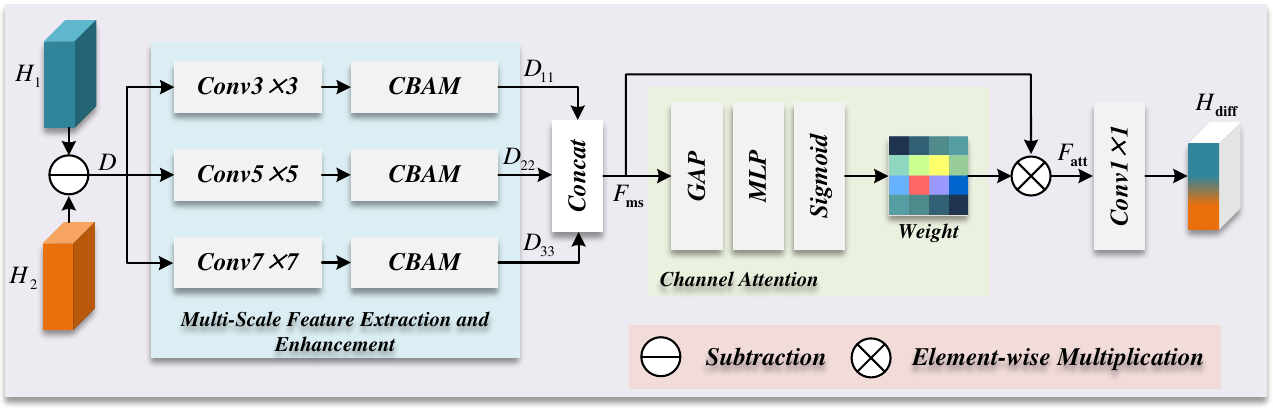}
\caption{The detailed structure of the FDID module.}
\label{figure}
\end{figure}
information under different receptive fields, the initial difference feature D is fed into three parallel convolutional branches with kernel sizes of $3\times3$, $5\times5$ and $7\times7$, respectively. Subsequently, the outputs of each branch are enhanced using a Convolutional Block Attention Module (CBAM), yielding multi-scale feature representations $D_{11}$, $D_{22}$, $D_{33}$. These multi-scale features are then concatenated along the channel dimension to form a fused feature representation $F_{ms}$. The CBAM module effectively highlights change regions and suppresses background noise interference by modeling spatial and channel dependencies. The above process can be formulated as follows:
\begin
{equation}
D=\parallel H_{1}-H_{2}\parallel,
\end
{equation} 
\begin
{equation}
D_{ii}=CBAM(Conv_{m\times m}(D)) (i=1,2,3) (m=3,5,7),
\end
{equation} 
\begin
{equation}
\mathrm{F_{ms}=Concat(D_{11},D_{22},D_{33})},
\end
{equation} 
where $\left\|\cdot\right\|$ denotes the absolute value operation.

To enhance key change regions, the FDID module introduces channel attention. Global average pooling (GAP) is applied to the fused features $F_{ms}$, followed by a two-layer MLP and Sigmoid activation to obtain attention weights. These weights re-emphasize salient features via element-wise multiplication. The refined output $F_{att}$ is then processed by a $1\times1$ convolution to generate high-frequency difference features $H_{diff}$, enabling precise modeling of subtle changes. The process is formulated as:
\begin
{equation}
F_{att}=F_{ms}\odot \sigma(MLP(GAP(F_{ms}))),
\end
{equation} 
\begin
{equation}
H_{diff}=Conv_{1\times1}(F_{att}),
\end
{equation} 
where $\sigma$ denotes the Sigmoid operation and $\odot$ represents the element-wise Hadamard product.

The proposed FDID module efficiently captures subtle changes in bi-temporal remote sensing images by integrating multi-scale features, CBAM attention, and high-frequency difference computation, while suppressing complex background interference. Its lightweight design facilitates easy integration, making it an effective and accurate component for change detection.

\subsection{Progressive Contextual Difference Module (PCDM)}

In remote sensing change detection, low-frequency features capture global semantics and are key to modeling bi-temporal differences. However, complex backgrounds (e.g., vegetation, roads) can obscure change information, reducing accuracy. To address this, we propose the Progressive Contextual Difference Module (PCDM) as shown in Fig. 3, which captures semantic changes from the low-frequency Transformer (TR) block and enhances global background modeling.

Specifically, the PCDM takes the bi-temporal low-frequency features $F_1^L$ and $F_2^L$ as input and performs pixel-wise differencing operations. Then, convolutional operations are applied to further enhance the difference features $D_i$, 
\begin{table*}[htbp]
\centering
\renewcommand{\arraystretch}{1.4} 
\setlength{\tabcolsep}{1pt} 
\caption{The comparison results of experiments on three datasets, with the highest values highlighted in \textcolor{red}{Red} and the second-highest values highlighted in \textcolor{blue}{blue}.}
\scriptsize 
\resizebox{0.965\textwidth}{!}{ 
\large 
\begin{tabular}{llcccccccccccc}
\toprule
\midrule
\multirow{2}{*}{\textbf{Datasets}} & \multirow{2}{*}{\textbf{Metric}} & \textbf{BIT} & \textbf{HFA-Net} & \textbf{ChangeFormer} & \textbf{FTN} & \textbf{ICIF-Net} & \textbf{AERNet} & \textbf{ELGC-Net} & \textbf{SEIFNet} & \textbf{CF-GCN} & \textbf{ChangeBind} & \multirow{2}{*}{\textbf{Ours}} \\ 
& & (TGRS$\cdot$22) & (PR$\cdot$22) & (IGARSS$\cdot$22) & (ACCV$\cdot$22) & (TGRS$\cdot$22) & (TGRS$\cdot$23) & (TGRS$\cdot$24) & (TGRS$\cdot$24) & (TGRS$\cdot$24) & (IGARSS$\cdot$24)\\ 
\midrule
\midrule
\multirow{6}{*}{\textbf{LEVIR}} 
& \textbf{OA} & 99.03 & 98.97 & 99.04 & 99.06 & 99.12 & 99.07 & 99.10 & 99.04 & 99.06 & {\color[HTML]{0000FF} 99.16} & {\color[HTML]{FF0000} 99.21}\\
& \textbf{F1} &90.34 &90.12 &90.40 &91.01 &91.18 &90.78 &91.06 &90.49 &91.07 &{\color[HTML]{0000FF} 91.41} & {\color[HTML]{FF0000} 91.97}\\
& \textbf{IoU} & 82.39 &81.70 &82.48 &83.51 &83.85 &83.11 &83.59 &82.62 &83.61 &{\color[HTML]{0000FF} 84.18} & {\color[HTML]{FF0000} 85.42}\\
& \textbf{Precision} &91.53 &92.27 &92.05 &92.71 &91.13 &89.97 &92.03 &91.52 &92.70 &{\color[HTML]{FF0000} 94.93} & {\color[HTML]{0000FF} 92.78}\\
& \textbf{Recall} &89.19 &87.69 &88.81 &89.37 &90.57 &{\color[HTML]{0000FF} 91.59} &90.11 &89.48 &89.20 &88.14 &{\color[HTML]{FF0000} 91.63}\\
& \textbf{Edge$_{\text{mIoU}}$} & 64.97 & 65.01 & 67.81 & 68.18 & 65.64 & 66.80 &68.17 & 65.88 & 61.63
& {\color[HTML]{0000FF}68.94} & {\color[HTML]{FF0000}72.31}\\
\midrule
\midrule
\multirow{6}{*}{\textbf{WHU}} 
& \textbf{OA} &98.81 &99.03 &98.76 &{\color[HTML]{FF0000} 99.37} &99.13 &99.25 &99.17 &99.04 &99.24 &99.28 &{\color[HTML]{0000FF} 99.34}\\
& \textbf{F1} & 87.47 &89.71 &86.88 &92.16 &90.77 &92.18 &91.18 &89.99 &91.97 &{\color[HTML]{0000FF} 92.31} &{\color[HTML]{FF0000} 93.37}\\
& \textbf{IoU} & 77.73 &78.53 &76.81 &85.45 &83.09 &85.49 &83.80 &81.80 &85.13 &{\color[HTML]{0000FF} 85.72} &{\color[HTML]{FF0000} 86.45}\\
& \textbf{Precision} &88.71 &91.74 &88.50 &93.09 &92.93 &92.47 &93.43 &90.11 &93.91 &{\color[HTML]{FF0000} 94.81} &{\color[HTML]{0000FF} 94.27}\\
& \textbf{Recall} & 86.27 &86.73 &85.33 &91.24 &88.70 &{\color[HTML]{0000FF} 91.89} &89.04 &89.88 &90.10 &89.94 &{\color[HTML]{FF0000} 91.96}\\
& \textbf{Edge$_{\text{mIoU}}$} & 68.16 & 68.43 & 67.69 & 71.21 & 69.43 & 70.98 & 71.08 & 69.12 & 70.05 & {\color[HTML]{0000FF}71.22} & {\color[HTML]{FF0000}73.39}\\
\midrule
\midrule
\multirow{6}{*}{\textbf{GZ}} 
& \textbf{OA} & 96.31 &96.99 &95.53 &{\color[HTML]{0000FF} 97.92} &97.29 &97.13 &96.85 &97.26 &97.08 &97.21 &{\color[HTML]{FF0000} 97.95}\\
& \textbf{F1} & 80.23 &82.75 &73.66 &{\color[HTML]{0000FF} 85.58} &85.09 &84.42 &82.67 &84.41 &83.65 &84.98 &{\color[HTML]{FF0000} 87.94}\\
& \textbf{IoU} & 66.99 &69.78 &58.30 &{\color[HTML]{0000FF} 74.79} &74.05 &73.03 &70.46 &73.62 &71.90 &73.89 &{\color[HTML]{FF0000} 78.83}\\
& \textbf{Precision} &82.40 &88.15 &84.59 &86.99 &89.90 &88.06 &87.24 &{\color[HTML]{0000FF} 90.45} &89.89 &87.81 &{\color[HTML]{FF0000} 93.19}\\
& \textbf{Recall} &78.18 &78.09 &65.23 &{\color[HTML]{0000FF} 84.21} &80.76 &81.07 &78.55 &79.82 &77.98 &82.34 &{\color[HTML]{FF0000} 84.32}\\
& \textbf{Edge$_{\text{mIoU}}$} & 56.17 & 56.42 & 51.02 & 58.92 & 58.47 & 58.04 & 59.06 & 58.85 & 57.96 & {\color[HTML]{0000FF}59.12} & {\color[HTML]{FF0000}61.04}\\
\midrule
\bottomrule
\end{tabular}}
\end{table*}
which are subsequently combined with the original low-frequency features through weighting to highlight the change regions. This process can be formulated as follows:
\begin
{equation}
D_{i}=Conv_{3\times3}(\parallel\mathrm{F_{1}^{L}-F_{2}^{L}\parallel}),
\end
{equation} 
\begin
{equation}
F_{j2} = (F_j^L \oplus D_i) \odot F_j^L,\quad j=1,2
\end
{equation} 
where $\left\|\cdot\right\|$ denotes the absolute value operation, $\odot$ represents the Hadamard product and $\oplus$ indicates element-wise addition.

To further capture finer-grained change information, we introduce multi-scale dilated convolutions with different dilation rates (7, 5, 3, 1) to extract features, obtaining temporal change features at various scales. This operation can be formalized as follows:
\begin
{equation}
F_{i4}=Conv_{3\times3}^{\alpha=7}(F_{i2})\oplus\mathrm{F_{i}^{L}},
\end
{equation} 
\begin
{equation}
F_{i6}=Conv_{3\times3}^{\alpha=5}(F_{i4})\oplus\mathrm{F_{i}^{L}},
\end
{equation} 
\begin
{equation}
F_{i7}=Conv_{3\times3}^{\alpha=3}(F_{i6}),
\end
{equation} 
\begin
{equation}
M_{i}^{L}=Conv_{3\times3}^{\alpha=1}(F_{i7}\oplus\mathrm{F_{i}^{L}}),
\end
{equation} 
where i=1, 2 . $\oplus$ represents element-wise addition.

Finally, we combine the fine-grained temporal difference representation with the coarse-grained difference representation to extract comprehensive and effective differential information. This operation can be formulated as follows:
\begin
{equation}
D^{LL}=Conv_{1\times1}(Conv_{3\times3}(C(M_1^L,M_2^L))\oplus D_i),
\end
{equation} 
where $C$ denotes the Concat operation, and $\oplus$ represents element-wise addition.

Through the above design, the PCDM effectively extracts temporal change features from dual-temporal images. By leveraging multi-scale convolutions and a progressive enhancement strategy, it precisely models the detailed information of the change regions. This module demonstrates enhanced robustness in complex backgrounds, accurately capturing subtle change areas while effectively suppressing the impact of background interference on the detection results.

\subsection{Loss Function}

In change detection tasks, there exists a severe imbalance between changed and unchanged samples, which often causes the model to be biased toward negative samples, thereby degrading detection performance. To address this issue, this letter adopts a combined loss function that integrates BCE loss and Dice loss, jointly optimizing model performance from both pixel-wise accuracy and region-level overlap perspectives. This enhances the model's sensitivity to change regions. The loss function is defined as follows:
\begin
{equation}
L_{BCE}(y,\hat{y})=-(y\log(\hat{y})+(1-y)\log(1-\hat{y})),
\end
{equation} 
\begin
{equation}
L_{Dice}(y,\hat{y})=1-\frac{2y\hat{y}+1}{y+\hat{y}+1},
\end
{equation} 
where $y$ and $\hat{y}$ represent the ground truth labels and the predicted labels, respectively.

The overall loss function is defined as follows:
\begin
{equation}
L_{total}(\mathrm{y},\hat{\mathrm{y}})=\lambda_1L_{BCE}(\mathrm{y},\hat{\mathrm{y}})+\lambda_2L_{Dice}(\mathrm{y},\hat{\mathrm{y}}),
\end
{equation} 
\begin{figure*}[htbp]
\centering
\includegraphics[scale=0.47]{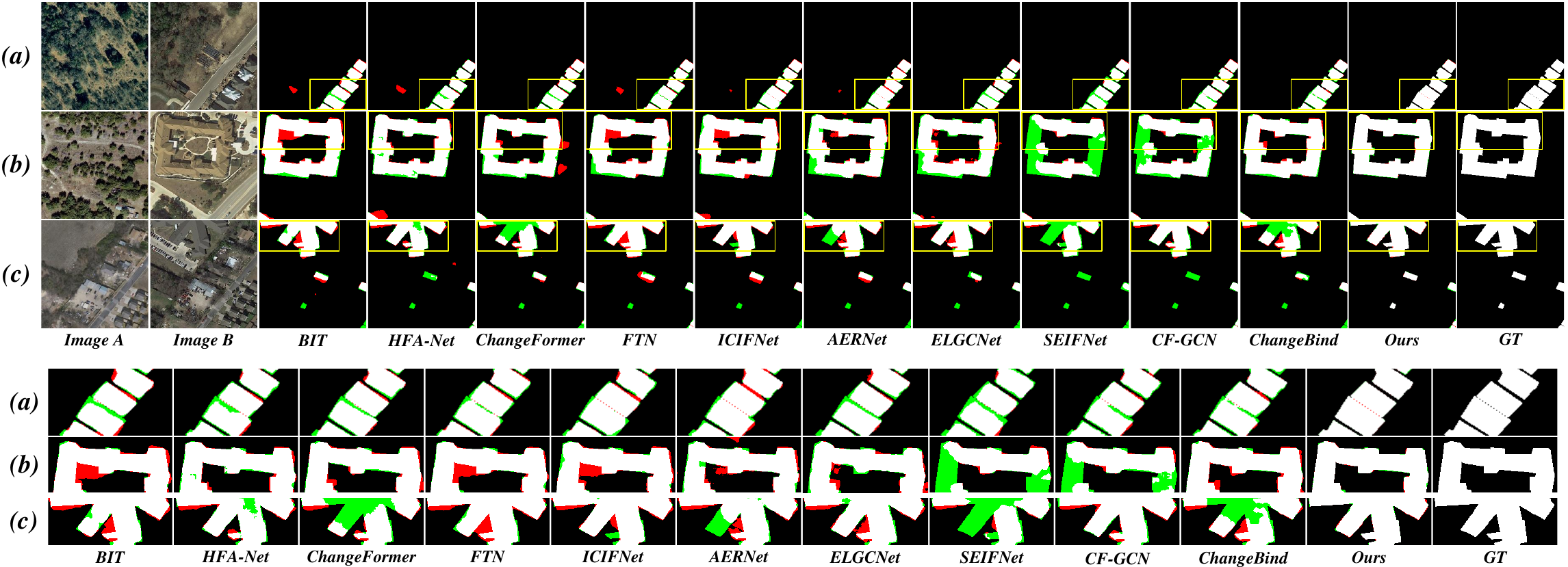}
\caption{Visualization results of our method compared to other state-of-the-art methods on the LEVIR-CD dataset. (a)-(c) show three randomly selected representative samples, where red indicates incorrectly identified regions and green represents unrecognized areas. Below, the zoomed-in regions of the yellow boxes from the three sample sets are displayed.}
\label{figure}
\end{figure*}
\begin{figure*}[htbp]
\centering
\includegraphics[scale=0.47]{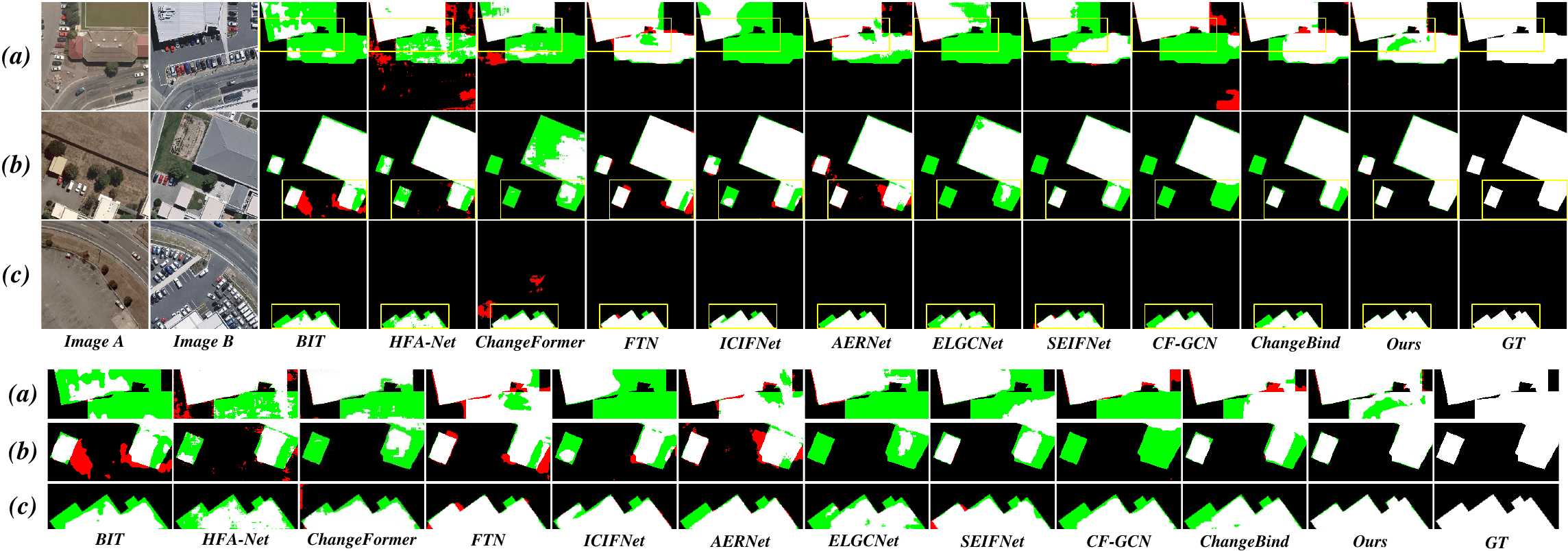}
\caption{Visualization results of our method compared to other advanced methods on the WHU-CD dataset. (a)-(c) show three randomly selected representative samples, where red indicates misclassified areas and green indicates undetected regions. Below, we display the zoomed-in views of the yellow boxed areas from the three samples.}
\label{figure}
\end{figure*}
where $\lambda_1$ and $\lambda_2$ are balancing parameters used to adjust the weights of the BCE loss and Dice loss, respectively. The appropriate weight parameters were determined to be 0.5 and 1 through experimental analysis.

\section{Experiment and evaluations}

\subsection{Datasets}

LEVIR-CD \cite{9} is a large-scale remote sensing dataset for building change detection. It contains a total of 31,333 image pairs of size $256\times256$, covering building additions and demolitions over a period of 5 to 14 years. The dataset is split into training, validation, and testing sets in a 7:1:2 ratio.

WHU-CD \cite{10} is a high-resolution remote sensing dataset designed for building change detection. It consists of two aerial images with a spatial resolution of 0.2 m, capturing building changes in Christchurch, New Zealand, following the 2011 earthquake. The data are cropped into $256\times256$ image patches and split into training, validation, and testing sets with a ratio of 8:1:1.

GZ-CD \cite{51} dataset contains high-resolution satellite image pairs from suburban Guangzhou, China, spanning 2006–2019. It includes 19 seasonal change pairs from Google Earth, with original sizes ranging from $1006\times11684$ to $4936\times5224$ pixels. All images are cropped into $256\times256$ patches and split into training, validation, and test sets at an 8:1:1 ratio, yielding 2504, 313, and 313 samples, respectively.

\subsection{Experimental Details}

WGDF is implemented with PyTorch 1.11 and trained on two NVIDIA TITAN RTX GPUs. Data augmentation (flipping, scaling, cropping, Gaussian blur) improves robustness and generalization. We adopt AdamW \cite{52} optimizer with momentum 0.9, weight decay 0.001, and $(\beta_1, \beta_2) = (0.99, 0.999)$. The model is trained for 300 epochs with a learning rate 
\begin{figure*}[htbp]
\centering
\includegraphics[scale=0.47]{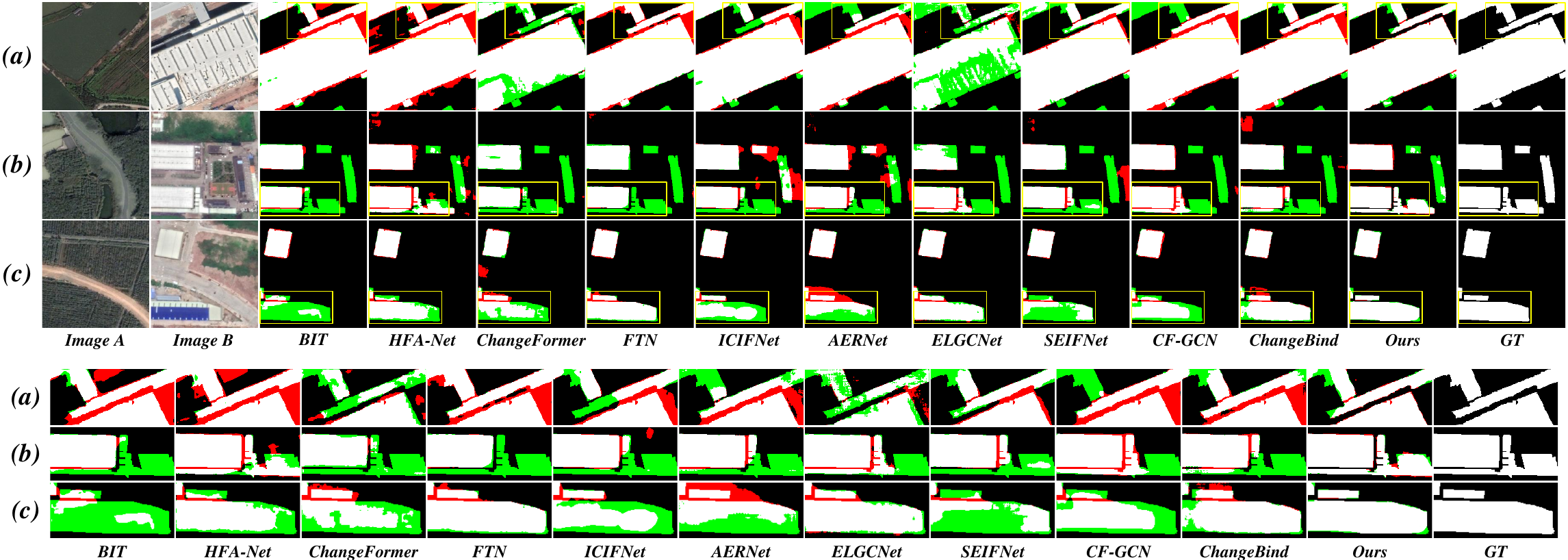}
\caption{Visualization results of our method compared to other advanced methods on the GZ-CD dataset. (a)-(c) show three randomly selected representative samples, where red indicates misclassified areas and green indicates undetected regions. Below, we display the zoomed-in views of the yellow boxed areas from the three samples.}
\label{figure}
\end{figure*}
of 0.0003 and batch size of 24, ensuring efficient convergence and performance.

Model performance is evaluated using five metrics: Precision, Recall, F1-Score, IoU, and Overall Accuracy (OA), which assess the similarity between predicted and ground truth change maps. 

To quantitatively evaluate edge detection performance, we use Edge mIoU, which measures the Intersection over Union (IoU) between predicted and ground truth change region edges. Edge mIoU is the average IoU computed over multiple change regions, calculated as follows:
\begin
{equation}
\mathrm{Edge~mIoU}=\frac{1}{N}\sum_{i=1}^N\mathrm{Edge~IoU}_i,
\end
{equation}
\begin
{equation}
\mathrm{Edge~IoU}_i=\frac{|E_p^i\cap E_g^i|}{|E_p^i\cup E_g^i|},
\end
{equation}
where $N$ represents the total number of change regions in the image, $E_p^i$ denotes the predicted change region edges, $E_g^i$ represents the ground truth change region edges, $|E_p^i\cup E_g^i|$ is the union of the predicted and ground truth edges, and $|E_p^i\cap E_g^i|$ is the intersection of the predicted and ground truth edges.

\subsection{Comparison with the State-of-the-Art Methods}

To comprehensively evaluate WGDF, we conducted extensive experiments on three dual-temporal remote sensing change detection datasets, comparing it against ten state-of-the-art methods: BIT \cite{17}, HFA-Net \cite{38}, ChangeFormer \cite{30}, FTN \cite{51}, ICIF-Net \cite{31}, AERNet \cite{53}, ELGC-Net \cite{34}, SEIFNet \cite{54}, CF-GCN \cite{55}, and ChangeBind \cite{56}. The experimental setup and comparisons were carefully designed to ensure rigorous evaluation.

\subsubsection{Experimental Results on LEVIR-CD} Table I reports quantitative results on the LEVIR-CD dataset. WGDF achieves the highest F1-score and IoU, outperforming ChangeBind by 0.56 and 1.24 points, respectively. Compared with BIT and HFA-Net, F1-score improves by 1.63 and 1.85 points, and IoU by 3.03 and 3.72 points. WGDF also shows strong Recall, surpassing AERNet and ELGC-Net by 0.04 and 1.52 points, respectively. Although Precision is 2.15 points lower than ChangeBind, it remains competitive. Notably, WGDF achieves the best Edge mIoU, excelling in boundary delineation and high-frequency detail preservation, especially around building edges and road intersections.

Fig. 5 presents three representative samples from the LEVIR-CD dataset, covering complex backgrounds with vegetation, roads, and buildings. These visualizations demonstrate WGDF's superior edge handling and change localization. In the first sample, our method more accurately distinguishes building edges from vegetation, reducing misdetections seen in other methods. The second sample shows improved robustness to occlusion and subtle changes, minimizing false detections. In the third, WGDF excels in low-contrast, small-scale changes, with results closely matching the ground truth. Zoomed-in views further highlight WGDF’s precise edge delineation, enabled by its effective use of high-frequency information in challenging scenarios.

\subsubsection{Experimental Results on WHU-CD} Table I shows WGDF’s results on the WHU-CD dataset, achieving top F1-score (93.37), IoU (86.45), and Recall (91.96), with second-best Precision (94.27). WGDF excels in Edge mIoU, accurately delineating boundaries in complex, detail-rich scenes. This demonstrates its effectiveness in reducing false detections and preserving fine edge details for reliable change detection.

Fig. 6 shows three representative WHU-CD samples covering complex scenarios like building edges, intersections, and low-contrast regions. Traditional methods (HFA-Net, BIT) miss edge details and produce errors. ChangeFormer and FTN improve but still suffer pseudo-changes. Our method accurately detects changes and localizes edges. In low-contrast and occluded areas, AERNet and ELGC-Net miss subtle changes or produce false positives, while ours is more robust. For small-scale changes, SEIFNet and CF-GCN struggle, but our results closely match ground truth with better 
\begin{figure*}[htbp]
\centering
\includegraphics[scale=0.433]{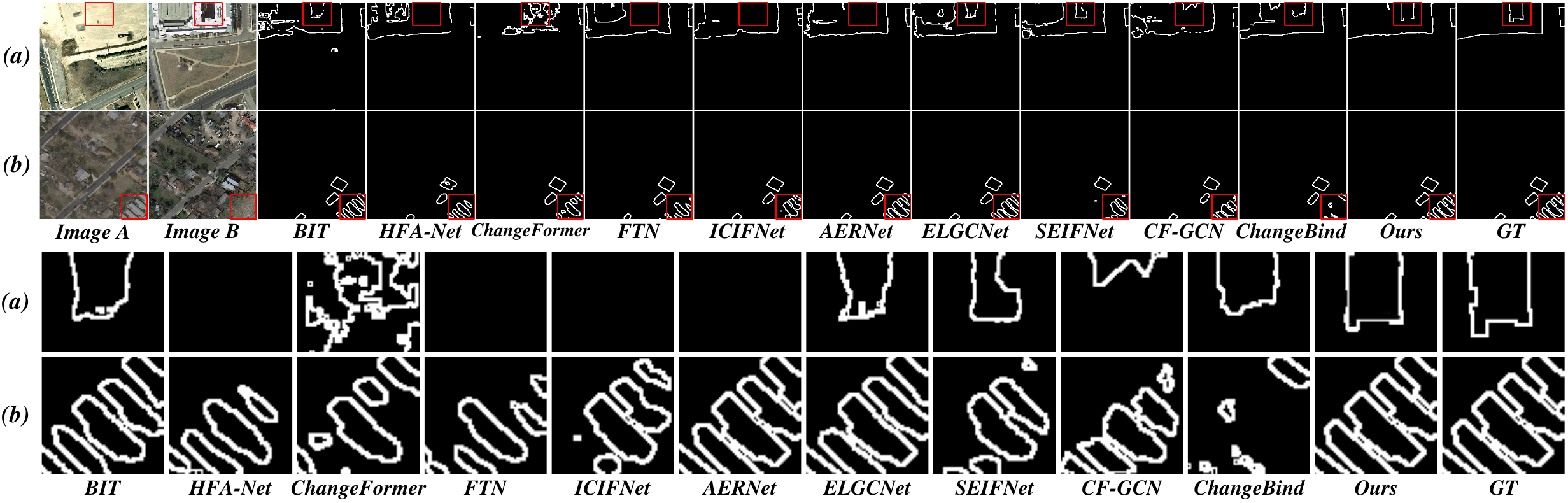}
\caption{Comparative Visualization of Fine-Grained Change Detection Results from Different Methods on the LEVIR-CD Dataset, with edge results generated by the Sobel operator. The lower part shows the zoomed-in detail area of the red-boxed region.}
\label{figure}
\end{figure*}
completeness and edge detail. Zoomed-in views highlight our superior edge restoration and precise localization in challenging backgrounds.

\subsubsection{Experimental Results on GZ-CD} Table I shows our method’s quantitative results on the GZ-CD dataset, achieving the highest scores in F1 (87.94) and IoU (78.83). Compared to the second-best FTN, our F1 and IoU improve by 2.36 and 4.04 points, respectively, demonstrating better accuracy and coverage. Against ChangeBind, improvements of 3.13 (F1) and 5.44 (IoU) further highlight our advantage in fine-grained detection. Precision (93.19) is significantly higher than FTN by 6.2 points, while Recall (84.32) is slightly lower by 0.11. Our method also leads in Edge mIoU, showing superior boundary delineation in complex regions, resulting in clearer, more reliable change detection.

In the GZ-CD dataset, three representative sample sets are visualized in Fig. 7. The first set, a dense building area, shows that HFA-Net and BIT suffer from many false detections due to road textures, while ChangeFormer and FTN miss some building details. Our method accurately detects building changes and effectively suppresses false positives. The second set involves vegetation and water boundaries with complex lighting and textures; ICIF-Net and AERNet produce more false positives, whereas our model accurately captures real changes and suppresses pseudo-changes. The third set covers road intersections and fine building changes with many small edges; SEIFNet and CF-GCN miss details or blur boundaries, but our method restores edges clearly and maintains high completeness. Zoomed-in views highlight that competing methods blur or partially miss edges, while ours precisely recovers boundary details and locates change regions.

\subsection{Validation of Change Detection in Fine-Grained Change Regions}

To further validate our model’s performance in fine-grained change regions, we analyzed representative samples using the Sobel operator for edge extraction. This evaluation focused on edge detail restoration and change region localization, with a comprehensive comparison against state-of-the-art methods.

Table I shows the edge mIoU results on the LEVIR-CD dataset. Our method achieves 72.31, outperforming the second-best ChangeBind (68.94) and significantly surpassing traditional methods like BIT (64.97). This highlights our model’s superior ability to restore edge details and accurately detect change regions in complex scenarios.

Fig. 8 illustrates the performance of various methods in detecting small-scale targets. HFA-Net and BIT often produce broken or missing edges in complex structures (e.g., Fig. 8(a)), while ChangeFormer, FTN, SEIFNet, and CF-GCN are prone to background noise in cluttered scenes (e.g., Fig. 8(b)), leading to missed detections of small buildings. In contrast, our method leverages DWT to decompose features into frequency components, enhancing high-frequency structural details and modeling ambiguous edges. This results in clearer boundaries, improved detection of small targets, and greater robustness to background interference.

\subsection{Ablation Experiments}

In order to fully validate the specific impact of each component on WGDF performance, we designed and implemented a series of ablation experiments. In the experiments, key modules were gradually removed or replaced to assess their contribution to the overall performance of the model, and the results of the ablation experiments were compared and analyzed in detail on three different datasets.

\subsubsection{Impact analysis of the number of DFFE and TR Blocks} To identify the optimal number of DFFE and TR blocks in the encoder, we conducted ablation studies comparing three 
\begin{figure*}[htbp]
\centering
\includegraphics[scale=0.41]{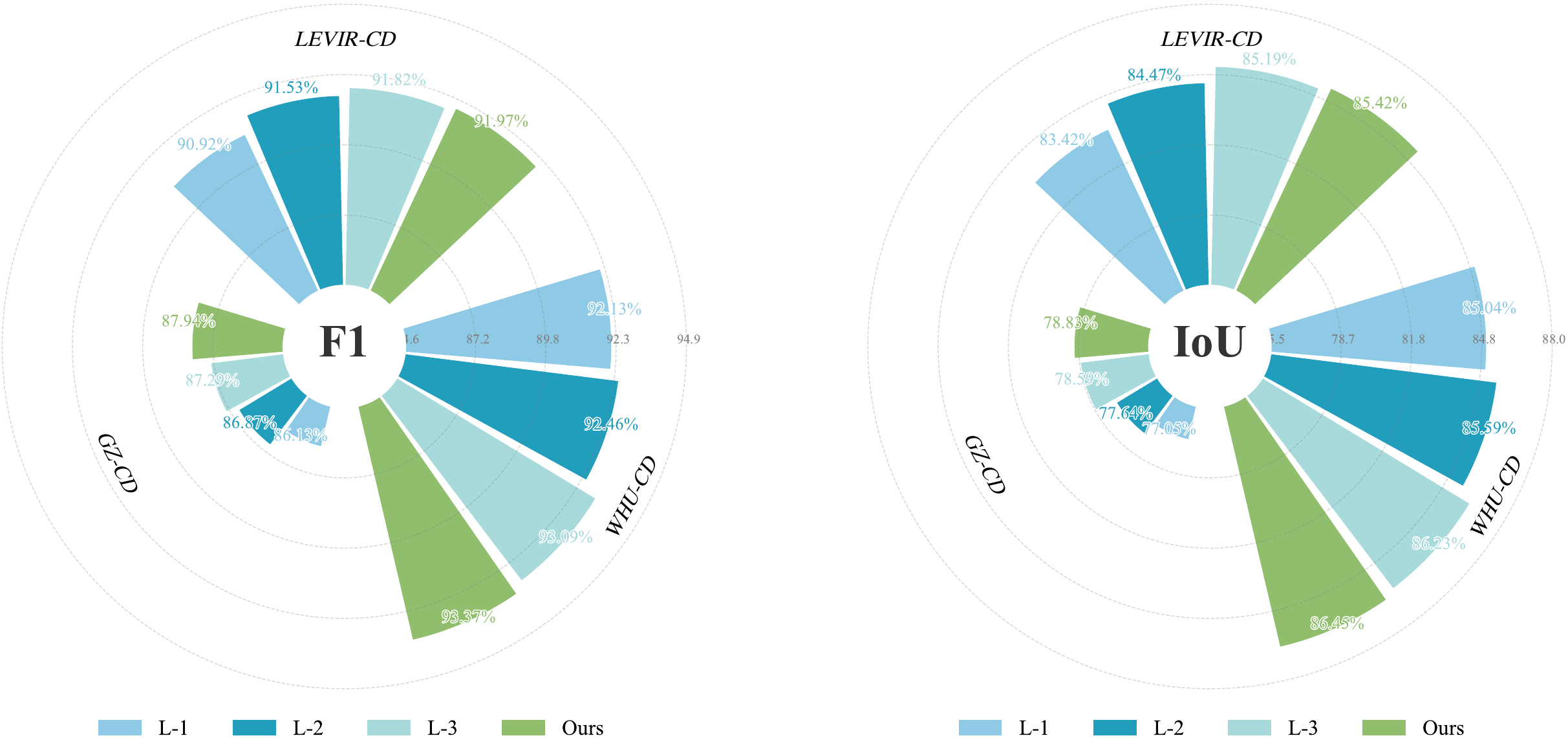}
\caption{Comparison of quantitative analysis results on three datasets with different loss function weights. Specifically, The parameter setting for L-1 is $\lambda_1$ = 0.2 and $\lambda_2$ = 0.5. The parameter setting for L-2 is $\lambda_1$ = 0.5 and $\lambda_2$ = 0.75. The parameter setting for L-3 is $\lambda_1$ = 0.6 and $\lambda_2$ = 0.4. The parameter setting for Ours is $\lambda_1$ = 0.5 and $\lambda_2$ = 1.}
\label{figure}
\end{figure*}
\begin{table}[htbp]
\centering
\renewcommand{\arraystretch}{1.4} 
\setlength{\tabcolsep}{6pt} 
\caption{Ablation experiments on the number of DFFE,TR block iterations on three datasets. Bold indicates the highest value.}
\begin{tabular}{lcccccc}
\toprule[1.1pt] 
\multirow{2}{*}{\textbf{Models}} & \multicolumn{2}{c}{\textbf{LEVIR-CD}} & \multicolumn{2}{c}{\textbf{WHU-CD}} & \multicolumn{2}{c}{\textbf{GZ-CD}} \\
\cline{2-7}
& \textbf{F1} & \textbf{IoU} & \textbf{F1} & \textbf{IoU} & \textbf{F1} & \textbf{IoU} \\
\toprule[1.1pt] 
\textbf{WGDF-2} & 91.36 & 84.19 & 92.74 & 85.59 & 86.47 & 78.52 \\
\textbf{WGDF-4} & 91.18 & 83.96 & 92.12 & 84.56 & 85.18 & 77.39 \\
\textbf{WGDF-3 (Ours)} & \textbf{91.97} & \textbf{85.42} & \textbf{93.37} & \textbf{86.45} & \textbf{87.94} & \textbf{78.83} \\
\toprule[1.1pt] 
\end{tabular}
\end{table}
\begin{table}[htbp]
\centering
\renewcommand{\arraystretch}{1.4} 
\setlength{\tabcolsep}{2.8pt} 
\caption{Ablation experiments on FDID and PCDM on three datasets. Bold indicates the highest value.}
\begin{tabular}{lcccccccc}
\toprule[1pt] 
\multirow{2}{*}{\textbf{Models}} & \multirow{2}{*}{\textbf{FDID}} & \multirow{2}{*}{\textbf{PCDM}} & \multicolumn{2}{c}{\textbf{LEVIR-CD}} & \multicolumn{2}{c}{\textbf{WHU-CD}} & \multicolumn{2}{c}{\textbf{GZ-CD}} \\
\cline{4-9}
& & & \textbf{F1} & \textbf{IoU} & \textbf{F1} & \textbf{IoU} & \textbf{F1} & \textbf{IoU} \\
\toprule[1.1pt] 
\textbf{Net-1} & × & × & 91.07 & 83.48 & 91.34 & 82.89 & 85.92 & 74.56 \\
\textbf{Net-2} & \checkmark & × & 91.71 & 84.83 & 92.84 & 85.97 & 87.89 & 77.46 \\
\textbf{Net-3} & × & \checkmark & 91.63 & 84.56 & 92.49 & 85.32 & 86.72 & 76.36 \\
\textbf{WGDF (Ours)} & \checkmark & \checkmark & \textbf{91.97} & \textbf{85.42} & \textbf{93.37} & \textbf{86.45} & \textbf{87.94} & \textbf{78.83} \\
\toprule[1.1pt] 
\end{tabular}
\end{table}
configurations. WGDF-2 uses two DFFE modules and two TR blocks, while WGDF-4 employs four of each. Our proposed WGDF-3 adopts three DFFE modules and three TR blocks, achieving the best balance between high- and low-frequency feature extraction. This configuration leads to improved generalization and robustness, outperforming the other variants across all tested datasets.

As shown in Table II, WGDF-2 performs relatively poorly due to insufficient DFFE and TR blocks, limiting effective extraction of high- and low-frequency features. Although WGDF-4 uses more modules, this can cause feature redundancy or training instability, slightly reducing performance. WGDF-3 (Ours) achieves the 
\begin{table}[htbp]
\centering
\renewcommand{\arraystretch}{1.4} 
\setlength{\tabcolsep}{6.7pt} 
\caption{Ablation experiments for the loss function. Bold indicates the highest value.}
\begin{tabular}{lcccccccc}
\toprule[1.1pt] 
\multirow{2}{*}{\textbf{Loss Function}} & \multicolumn{2}{c}{\textbf{LEVIR-CD}} & \multicolumn{2}{c}{\textbf{WHU-CD}} & \multicolumn{2}{c}{\textbf{GZ-CD}} \\
\cline{2-7}
& \textbf{F1} & \textbf{IoU} & \textbf{F1} & \textbf{IoU} & \textbf{F1} & \textbf{IoU} \\
\toprule[1.1pt] 
BCE & 91.76 & 84.68 & 92.78 & 85.92 & 87.96 & 77.43 \\
Dice & 91.59 & 84.51 & 92.81 & 85.96 & 86.29 & 75.28 \\
BCE+Dice & \textbf{91.97} & \textbf{85.42} & \textbf{93.37} & \textbf{86.45} & \textbf{87.94} & \textbf{78.83} \\
\toprule[1.1pt] 
\end{tabular}
\end{table}
best results across all three datasets, demonstrating that three pairs of DFFE and TR blocks provide an optimal balance of frequency features, enhancing generalization and robustness.

\subsubsection{Ablation Experiment of FDID} To validate the role of the FDID module in high-frequency feature extraction, we replaced it with a simple feature subtraction operation and conducted ablation studies (Table III). Results show that removing FDID (Net-3) leads to notable performance drops across all datasets, especially on GZ-CD, where F1 and IoU drop by 1.22 and 2.47 points, respectively. In contrast, introducing FDID (Net-2) yields consistent improvements over the baseline (Net-1), particularly on complex scenes like GZ-CD. These results confirm that FDID's multi-scale feature interaction and enhancement effectively boost the model’s ability to capture fine-grained changes and suppress background interference.

\subsubsection{Ablation Experiments with PCDM} PCDM is designed to enhance low-frequency semantic representation by aggregating multi-scale context, improving the model’s ability to distinguish change regions from background. To verify its effectiveness, we replaced PCDM with simple feature subtraction (Net-2) and observed notable performance degradation across datasets (Table III), particularly on WHU-CD and GZ-CD. Conversely, introducing PCDM (Net-3) leads to improved F1 and IoU scores, demonstrating 
\begin{figure}[htbp]
\centering
\includegraphics[scale=0.33]{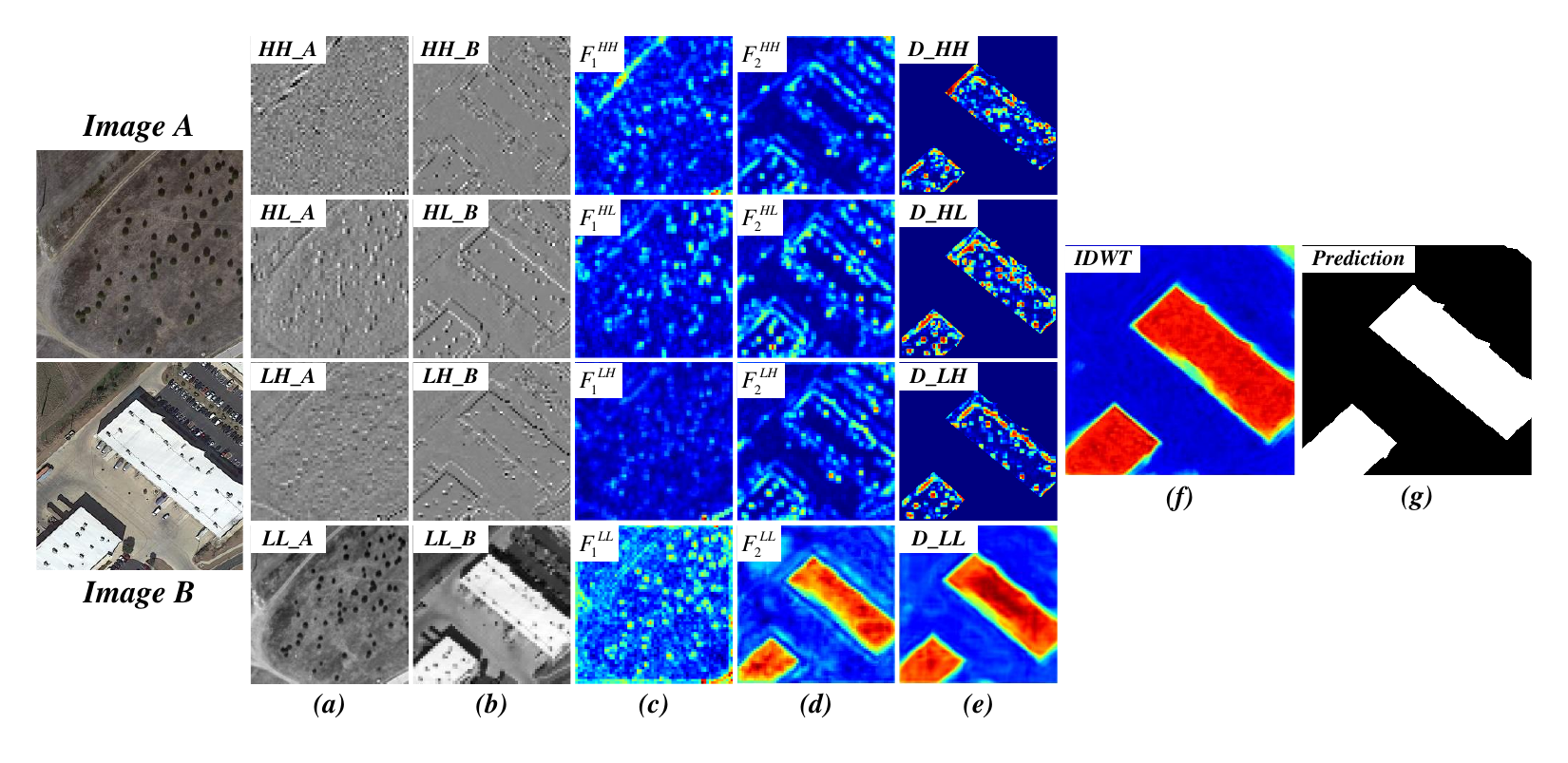}
\caption{This figure illustrates the WGDF workflow using LEVIR-CD samples. (a) and (b) show the DWT-decomposed subbands (HH, HL, LH, LL) of Images A and B. (c) and (d) depict the high- and low-frequency feature maps, representing multi-scale subband information. (e) presents the subband-wise difference features. (f) is the fused feature map reconstructed via IDWT. (g) shows the final change detection result.}
\label{figure}
\end{figure}
its effectiveness in modeling global semantics and suppressing background interference. These results highlight PCDM’s critical role in enhancing low-frequency semantic discrimination and overall robustness.

\subsubsection{Ablation Analysis of the Loss Function} To assess the impact of loss functions, we conducted ablation experiments (Table IV). Results show that using BCE or Dice loss alone leads to suboptimal performance on all datasets. In contrast, their combination consistently improves F1 and IoU scores—particularly on WHU-CD and GZ-CD—demonstrating enhanced ability to capture semantic differences and edge details. The joint loss leverages BCE’s classification precision and Dice’s robustness to class imbalance, effectively mitigating edge ambiguity and improving change detection accuracy in complex backgrounds.

\subsubsection{Ablation Analysis of Loss Function Weight Parameters} We conducted ablation experiments to analyze the effect of the loss weight parameters $\lambda_1$ and $\lambda_2$ on model performance, as shown in Fig. 9. By varying these values across the LEVIR-CD, WHU-CD, and GZ-CD datasets, we found that model accuracy is sensitive to their settings. The best performance in terms of F1 and IoU was achieved when $\lambda_1$ = 0.5 and $\lambda_2$ = 1, suggesting this combination provides an effective balance between classification accuracy and structural consistency.

\subsection{Parametric Analysis}

Table V compares different change detection methods in terms of model complexity (parameters, FLOPs, inference time) and performance (F1, IoU). WGDF achieves top performance with moderate complexity: 17.98M parameters—higher than BIT (3.55M) and CF-GCN (13.58M) but much smaller than FTN (168.53M) and ChangeFormer (41.03M). WGDF’s FLOPs (35.64G) are only 8.4\% of FTN’s (421.67G) and much lower than ChangeFormer’s (202.87G). On the WHU-CD dataset, WGDF’s inference time is 180.73s, 57.1\% and 51.6\% faster than FTN and ChangeFormer respectively. These results demonstrate WGDF’s efficient architecture that balances accuracy and computational cost, effectively capturing edge features and complex backgrounds. Future work will focus on further network refinement and lightweight designs for larger and more complex scenarios.

\subsection{Network Visualization}

To better illustrate WGDF’s architecture, Fig. 10 visualizes its key stages. (a) and (b) show the Discrete Wavelet Transform (DWT) band decomposition of input images A and B into four frequency bands (HH, HL, LH, LL). (c) and (d) display rich features extracted from high- and low-frequency branches that highlight change regions. (e) presents difference features enhancing change saliency. (f) shows feature fusion via Inverse DWT (IDWT), and (g) depicts the final change detection result.

\section{CONCLUSION}
\noindent 

This paper proposes Wavelet-Guided Dual-Frequency Encoding (WGDF), an efficient frequency-domain method to tackle edge ambiguity in remote sensing change detection. Input images are decomposed via Haar DWT into high- and low-frequency branches. The high-frequency branch employs the Dual-Frequency Feature Enhancement Module (DFFE) and Frequency-Domain Interactive Difference Module (FDID) to extract edge changes and suppress background noise. The low-frequency branch models global semantics with a Transformer block and refines change regions using the Progressive Contextual Difference Module (PCDM). A combined BCE and Dice loss enhances robustness to class imbalance. Experiments on three benchmark datasets demonstrate WGDF’s superior accuracy and efficiency, especially in complex scenes. Ablation studies confirm its robustness. WGDF advances frequency-domain feature modeling and balances detection accuracy with computational cost, offering theoretical insights for remote sensing and broader vision tasks.

\bibliographystyle{IEEEtran}
\bibliography{ref.bib}
\begin{IEEEbiography}[{\includegraphics[width=1in,height=1.25in,clip,keepaspectratio]{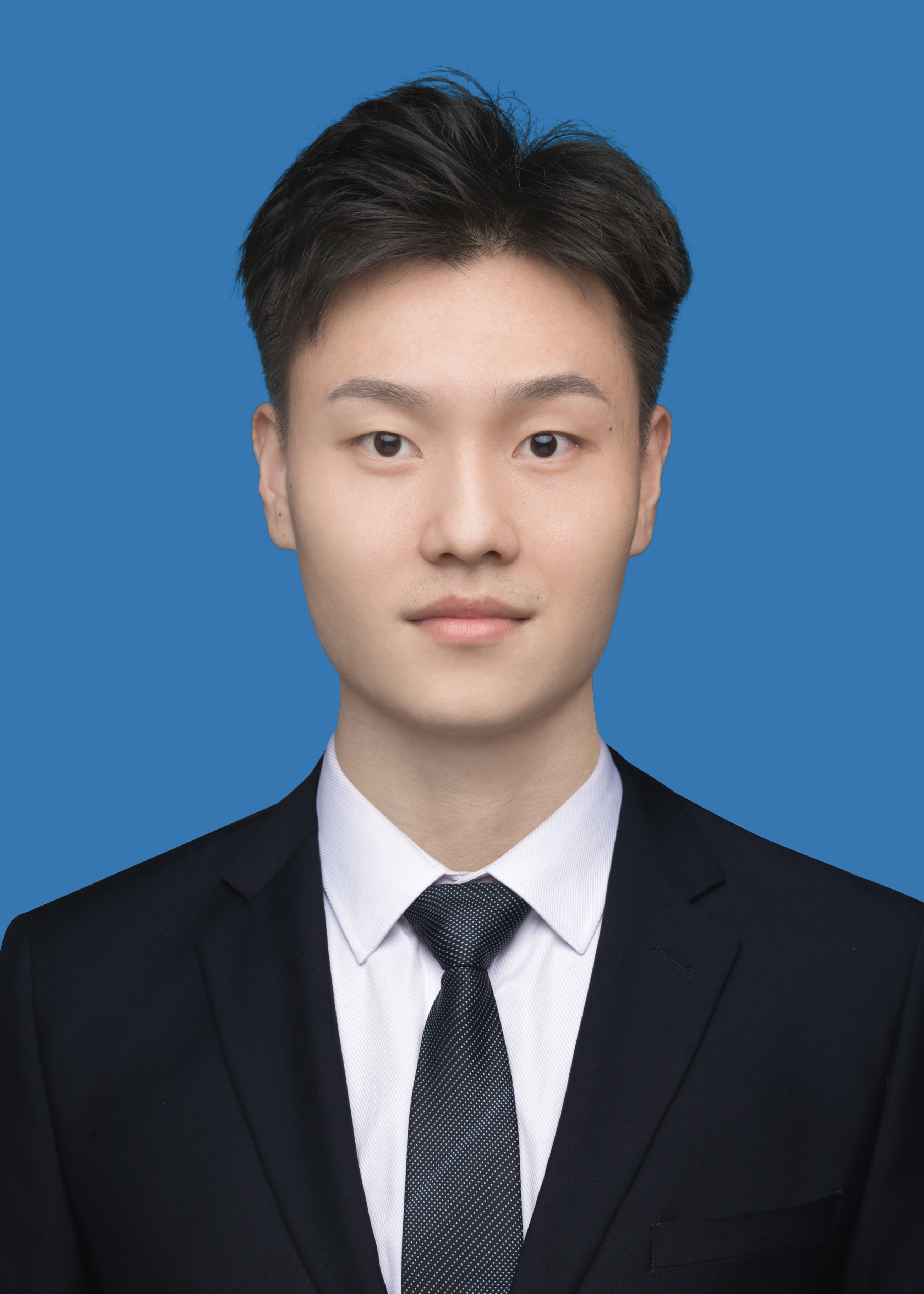}}]{Xiaoyang Zhang}
	graduated from Shandong Technology and Business University, School of Information and Electronic Engineering in 2023 with a bachelor's degree. Currently studying for a master’s degree in the School of Information and Electronic Engineering, Shandong Technology and Business University, Yantai, Shandong. His research interests include computer vision and image processing.
\end{IEEEbiography}


\begin{IEEEbiography}[{\includegraphics[width=1in,height=1.25in,clip,keepaspectratio]{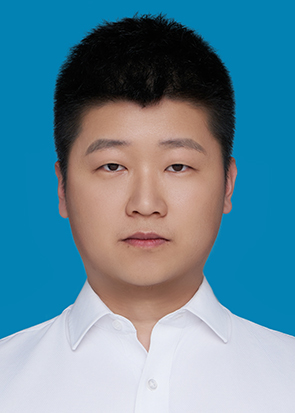}}]{Guodong Fan}
	received the M. Eng. degree from Shandong Technology and Business University, China, in 2021, the Ph.D. degree in Qingdao University, Qingdao, China, in 2025. He is currently a Associate Professor at the school of computer science and technology, Shandong Technology and Business University.  His research interests are in image processing, machine learning, and computer vision.
\end{IEEEbiography}
\vspace{-25pt}
\begin{IEEEbiography}[{\includegraphics[width=1in,height=1.25in,clip,keepaspectratio]{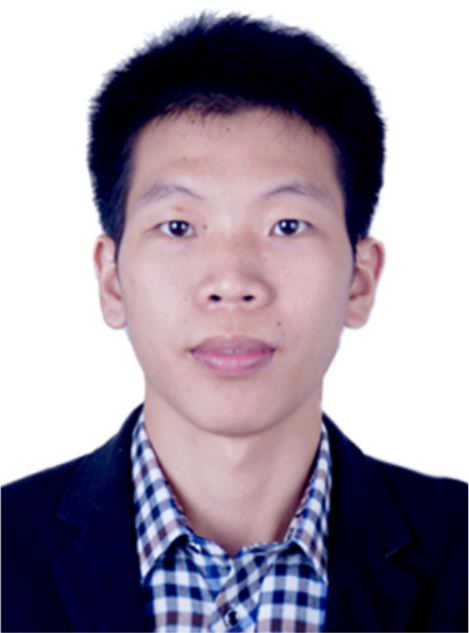}}]{Guang-Yong Chen}(Member IEEE) received the B.S. degree in mathematics from Xidian University, Xi'an, Shanxi 710071, China, in 2012, M. S. degree in mathematics from University of Science and Technology of China Hefei, Anhui, 230026, China in 2014, and the Ph.D. degree in mathematics from the College of Mathematics and Computer Science, Fuzhou University, Fuzhou, 350116, China. 
    
    He is currently a professorial fellow in the College of Computer and Data Science, Fuzhou University, Fuzhou, 350116, China. His research interests include computational intelligence, system identification, and nonlinear time series analysis.
\end{IEEEbiography}
\vspace{-25pt}
\begin{IEEEbiography}[{\includegraphics[width=1in,height=1.25in,clip,keepaspectratio]{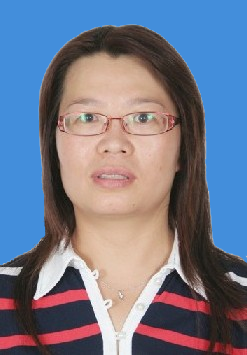}}]{Zhen Hua}
	received the B.S. and M.S. degrees in electrical automation from Taiyuan University of Technology, Taiyuan, China, in 1989 and 1992, respectively, the Ph.D. degree in electronic information engineering from China University of Mining and Technology, Beijing, China, in 2008. She is currently a professor at Shandong Technology and Business University. Her research interests include computer aided geometric design, information visualization, virtual reality, and image processing.
\end{IEEEbiography}
\vspace{-25pt}
\begin{IEEEbiography}[{\includegraphics[width=1in,height=1.25in,clip,keepaspectratio]{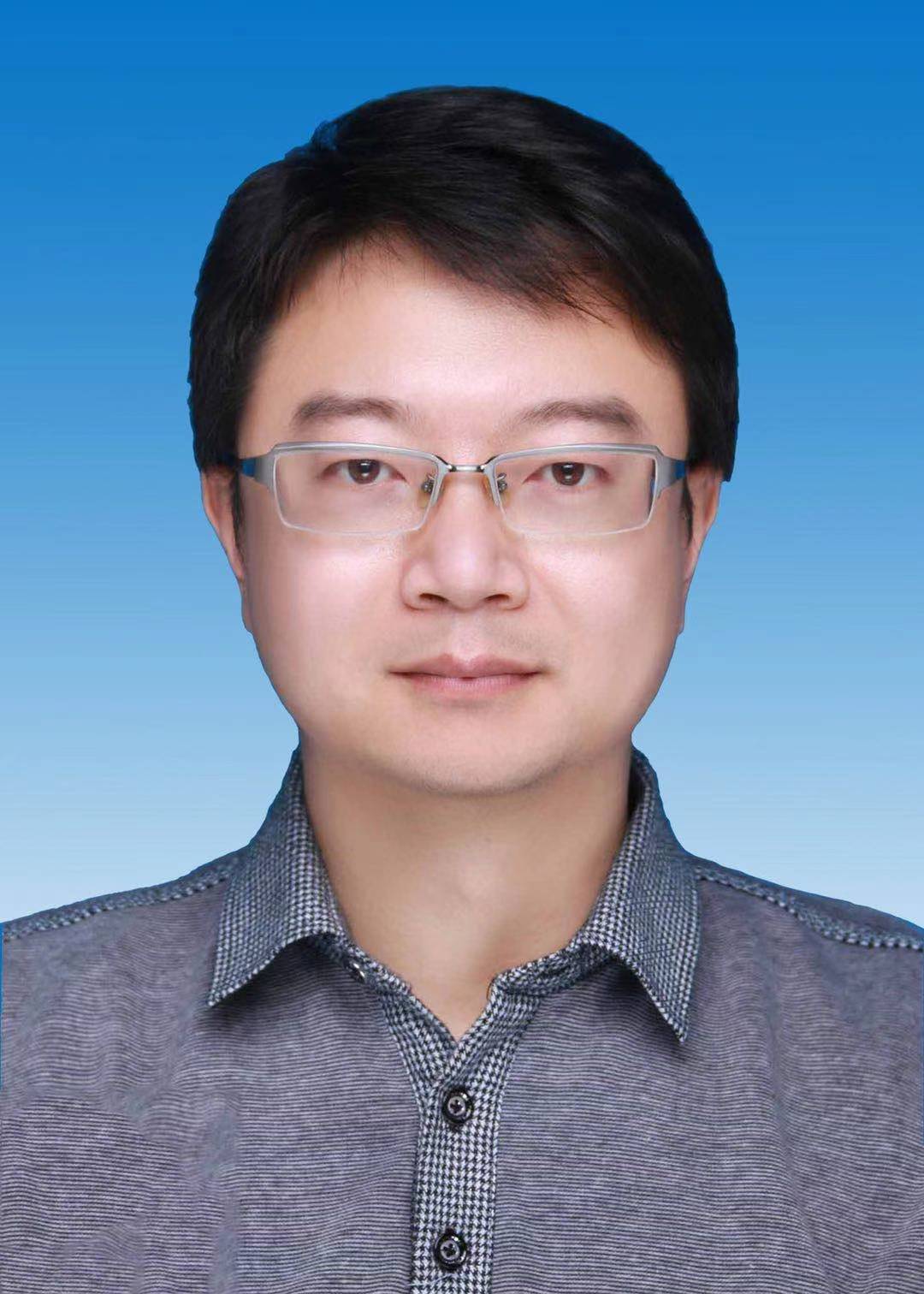}}]{Jinjiang Li}
	received the B. S. and M. S. degrees in computer science from Taiyuan University of Technology, Taiyuan, China, in 2001 and 2004, respectively, the Ph. D. degree in computer science from Shandong University, Jinan, China, in 2010. From 2004 to 2006, he was an assistant research fellow at the institute of computer science and technology of Peking University, Beijing, China. From 2012 to 2014, he was a Post-Doctoral Fellow at Tsinghua University, Beijing, China. He is currently a Professor at the school of computer science and technology, Shandong Technology and Business University. His research interests include image processing, computer graphics, computer vision, and machine learning.
\end{IEEEbiography}
\vspace{-25pt}
\begin{IEEEbiography}[{\includegraphics[width=1in,height=1.25in,clip]{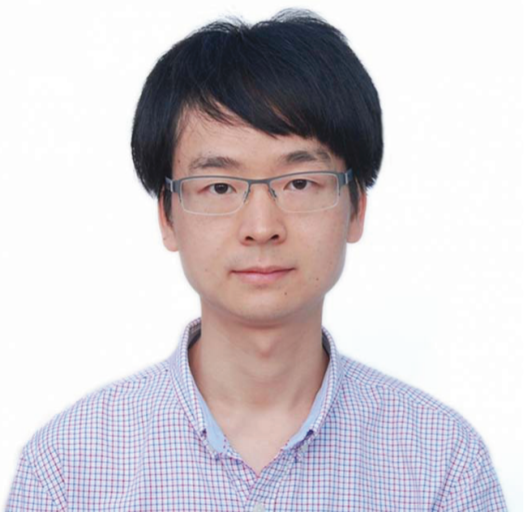}}]%
    {Min Gan} (Senior Member, IEEE) 
    received the B. S. degree in Computer Science and Engineering from Hubei University of Technology, Wuhan, China, in 2004, and the Ph.D. degree in Control Science and Engineering from Central South University, Changsha, China, in 2010. He is currently a professor in the College of Computer Science and Technology, Qingdao University, Qingdao, China. His current research interests include computer vision, statistical learning, system identification and nonlinear time series analysis.  
\end{IEEEbiography}
\vspace{-25pt}
\begin{IEEEbiography}[{\includegraphics[width=1in,height=1.35in,clip]{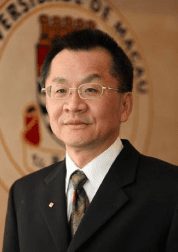}}]%
    {C. L. Philip Chen} (Life Fellow, IEEE) received the M.S. degree in electrical engineering from the University of Michigan, Ann Arbor, MI, USA, in 1985, and the Ph.D. degree in electrical engineering from Purdue University, West Lafayette, IN, USA, in 1988.
    
    He was a Tenured Professor, the Department Head, and an Associate Dean with two different universities in U.S. for 23 years. He is currently the Head of the School of Computer Science and Engineering, South China University of Technology, Guangdong, China. His current research interests include systems, cybernetics, and computational intelligence. He is currently the Dean of the School of Computer Science and Engineering, South China University of Technology, Guangzhou 510641, China. His current research interests include systems, cybernetics, and computational intelligence. 

    Dr. Chen received the 2016 Outstanding Electrical and Computer Engineers Award from his alma mater, Purdue University. He was the IEEE SMC Society President from 2012 to 2013 and the Vice President of Chinese Association of
    Automation. He has been the Editor-in-Chief of the IEEE TRANSACTION ON SYSTEMS, MAN, AND CYBERNETICS: SYSTEMS, since 2014 and an associate editor of several IEEE TRANSACTIONS. He was the Chair of TC 9.1 Economic and Business Systems of International Federation of Automatic Control (2015–2017), and also a Program Evaluator of the Accreditation Board of Engineering and Technology Education of the U.S. for Computer Engineering, Electrical Engineering, and Software Engineering Programs. He is a Fellow of AAAS, IAPR, CAA, and HKIE.
\end{IEEEbiography}

\end{document}